\documentclass[journal,draftclsnofoot,onecolumn,12pt,twoside]{IEEEtranTCOM}
\normalsize
\IEEEoverridecommandlockouts
\usepackage{xcolor}
\usepackage{tikz}
\usetikzlibrary{spy,arrows}
\usepackage{pgfplots}
\usepackage[ruled]{algorithm2e}
\SetKwInput{kwInit}{Initialization}
\usepackage{amsmath,epsfig,amssymb,amsthm,cite,url,xcolor}
\usepackage{hyperref}
\usepackage{enumerate}
\usepackage[latin1]{inputenc}

\theoremstyle{plain}
\newtheorem{theorem}{Theorem}
\newtheorem{assumption}{Assumption}
\newtheorem{lemma}[theorem]{Lemma}

\theoremstyle{definition}

\newtheoremstyle{specialcasestyle}{1mm}{1mm}{\upshape}{}{\bfseries\upshape}{.}{0mm}{}
\theoremstyle{specialcasestyle}

\newcommand{\figref}[1]{Fig.~\protect\ref{#1}}

\newcommand{\bu}{{\bf u}}

\newcommand{\bx}{{\bf x}}
\newcommand{\bv}{{\bf v}}

\newcommand{\by}{{\bf y}}
\newcommand{\bw}{{\bf w}}

\newcommand{\bz}{{\bf z}}

\newcommand{\bI}{{\bf I}}

\newcommand{\bm}{{\bf m}}

\newcommand{\ex}{{\mathbb E}}

\newcommand{\gm}{{\rm geomed}}

\newcommand{\bxi}{{\boldsymbol\xi}}

\makeatletter
\def\endthebibliography{%
  \def\@noitemerr{\@latex@warning{Empty `thebibliography' environment}}%
  \endlist
}
\makeatother
\usepackage{graphicx}
\usepackage{placeins}
\usepackage{float}
\usepackage{tabularx}
\usepackage{tkz-euclide,subfigure}
\usepackage{amsmath,epsfig,amssymb}
\usepackage{amsmath}
\usepackage{bbm}
\usetikzlibrary{intersections}
\usepgfplotslibrary{fillbetween}

\usepackage{mathtools}
\pgfplotsset{compat=1.12}
\usepackage[detect-all]{siunitx}
\usepackage{etoolbox}
\makeatletter
\patchcmd{\@makecaption}
  {\scshape}
  {}
  {}
  {}
\makeatother
\def\tablename{Table}

\DeclarePairedDelimiter{\norm}{\lVert}{\rVert} 
\begin{document}

\title{Over-The-Air Federated Learning under Byzantine Attacks}
\author{Houssem Sifaou,~\IEEEmembership{Member,~IEEE}, Geoffrey Ye Li,~\IEEEmembership{Fellow,~IEEE}
\thanks{} \thanks{}
\thanks{
Houssem Sifaou and Geoffrey Ye Li are with the Department of Electrical and Electronic Engineering, Imperial College London, London SW7 2AZ, U.K. (h.sifaou@imperial.ac.uk; geoffrey.li@imperial.ac.uk)}}

\maketitle

\begin{abstract}
Federated learning (FL) is a promising solution to enable many AI applications, where sensitive datasets from distributed clients are needed for collaboratively training a global model. FL allows the clients to participate in the training phase, governed by a central server, without sharing their local data. One of the main challenges of FL is the communication overhead, where the model updates of the participating clients are sent to the central server at each global training round. Over-the-air computation (AirComp) has been recently proposed to alleviate the communication bottleneck where the model updates are sent simultaneously over the multiple-access channel. However, simple averaging of the model updates via AirComp makes the learning process vulnerable to random or intended modifications of the local model updates of some Byzantine clients. In this paper, we propose a transmission and aggregation framework to reduce the effect of such attacks while preserving the benefits of AirComp for FL. For the proposed robust approach, the central server divides the participating clients randomly into groups and allocates a transmission time slot for each group. The updates of the different groups are then aggregated using a robust aggregation technique. We extend our approach to handle the case of non-i.i.d. local data, where a resampling step is added before robust aggregation. We analyze the convergence of the proposed approach for both cases of i.i.d. and non-i.i.d. data and demonstrate that the proposed algorithm converges at a linear rate to a neighborhood of the optimal solution. Experiments on real datasets are provided to confirm the robustness of the proposed approach.  
%This paper investigates the robustness of over-the-air federated learning to Byzantine attacks. The simple averaging of the model updates via over-the-air computation makes the learning task vulnerable to random or intended modifications of the local model updates of some malicious clients. We propose a robust transmission and aggregation framework to such attacks while preserving the benefits of over-the-air computation for federated learning. For the proposed robust federated learning, the participating clients are randomly divided into groups and a transmission time slot is allocated to each group. The parameter server aggregates the results of the different groups using a robust aggregation technique and conveys the result to the clients for another training round. We also analyze the convergence of the proposed algorithm. Numerical simulations confirm the robustness of the proposed approach to Byzantine attacks. 
\end{abstract}

\section{Introduction}

With the rapid growth and unprecedented success of artificial intelligence (AI) applications, a huge amount of data is transferred every day from distributed clients to data centers for further processing. However, in many applications, the data is sensitive, and sending it to data centers constitutes a major privacy concern \cite{10.1145/335191.335438,duchi2013local,Zhou2018}. Federated learning (FL) has emerged as a promising solution to such privacy concern\cite{pmlr-v54-mcmahan17a,DBLP:journals/corr/abs-1912-04977,9530714}. With FL, multiple distributed clients train a global machine learning model without sharing their local data. Local computations are carried out at the different clients and only the model parameter updates (or gradients) are sent to the central server (CS). The latter aggregates the local updates and forwards the result to the distributed clients for another training round.

Although data privacy is preserved using FL, allowing the clients to perform model parameter update opens the way for possible adversarial attacks. Indeed, in some cases, some of the distributed clients may send modified parameter updates with the intention of misleading the learning process \cite{6582732,8454396}. In this context, Byzantine attacks are a popular class, where certain clients aim to prevent the model from converging or causing convergence to a falsified model. The Byzantine clients may act independently or collectively. Unfortunately, even a single malicious client in a distributed setup as in FL can seriously affect the end model \cite{blanchard2017machine}. Developing countermeasures to these attacks has gained an increasing interest in recent years. Considering the FL setup without communication constraint, several aggregation techniques have been proposed to robustify the stochastic gradient descent (SGD) in the distributed setup. Stochastic algorithms tolerating few Byzantine attacks has been developed by aggregating the stochastic gradients updates using the median \cite{Xie2018GeneralizedBS}, geometric median\cite{minsker2015geometric,10.1145/3154503}, trimmed mean \cite{yin2018byzantine}, and iterative filtering \cite{10.1145/3322205.3311083}. The Krum aggregation in \cite{blanchard2017machine} selects the stochastic gradient having a minimum distance from a certain number of nearest stochastic gradients. A robust stochastic aggregation (RSA) has been developed in \cite{li2019rsa}, which tolerates heterogeneous datasets and Byzantine attacks. Other related works include leveraging redundancy of gradients to enhance robustness \cite{chen2018draco,Rajput2019DETOXAR} and avoiding saddle points in non-convex distributed learning in presence of  Byzantine attacks \cite{yin2019defending}. Moreover, the advantages of reducing the variance of the stochastic gradients to defend against Byzantine attackers have been investigated in \cite{9153949,peng2020byzantine}. 

All the above works are applicable when the individual local updates are sent separately to the CS. Sending the individual model updates at each training round constitutes a significant communication overhead especially for wireless devices. This becomes more and more significant when the number of participating clients is huge which is the case of most real world applications of FL. To alleviate this communication overhead, over-the-air computation has been proposed as a potential solution, where the local model updates are sent simultaneously over the multiple-access channel \cite{Kai2020,9042352}. Such approach can reduce significantly the communication bottleneck of FL \cite{9014530,Sery2020}. Many works have investigated the convergence of over-the-air FL (OTA-FL). However, its robustness to Byzantine attacks has not been well investigated. %Indeed, all the aforementioned Byzantine defenses are not applicable in the case of OTA-FL. 

This paper is one of the first attempts to tackle the Byzantine attacks in case of OTA-FL. To the best of our knowledge, the only works that tried addressing the Byzantine attacks problem in OTA-FL are \cite{9473694,fan2022bevsgd}. While  the Weiszfeld algorithm for geometric aggregation using AirComp has been implemented in \cite{9473694}, a best effort voting (BEV) power control policy for OTA SGD has been proposed in \cite{fan2022bevsgd} to defend against Byzantine attacks by letting the clients transmit with their maximum power.

%However, in the case of over-the-air FL (OTA-FL) \cite{Kai2020,9042352}, the local model updates are sent simultaneously over the analog wireless channel. This makes the aforementioned robust aggregation techniques not directly applicable in the case of OTA-FL.

In this work, to accommodate Byzantine attacks in the case of OTA-FL, we propose a transmission and aggregation approach that simultaneously exploits the benefits of over-the-air computation while being robust to Byzantine attacks, which is named 'ROTAF' standing for robust OTA-FL. Our approach consists of dividing the participating clients into several groups at each global training round, assigning a transmission time slot for each group, and aggregating the model updates of different groups using geometric median aggregation. The convergence of the proposed approach is analyzed under some assumptions on the loss functions. Based on our analysis, when the number of attacks is less than half of the number of groups, the proposed algorithm converges at a linear rate to a neighborhood of the optimal solution with an asymptotic learning error that depends on the noise variance and the number of Byzantine attackers. Moreover, as evidenced by numerical results, the proposed approach is robust to Byzantine attacks compared with the simple averaging OTA-FL. The approach is then extended to handle the case of non-i.i.d. data, where a simple resampling step is added before the geometric median aggregation. Such step can reduce significantly the variance of the group updates and thus enhance the performance of the geometric median aggregation.

The main contributions of the paper are summarized as follows:
\begin{itemize}
\item[1)] We propose a transmission and aggregation framework to deal with Byzantine attacks in OTA-FL. The proposed framework can handle both i.i.d. and non-i.i.d. data distributions.
\item[2)] We conduct a theoretical convergence analysis of the proposed framework for both i.i.d. and non-i.i.d. data distributions. Specifically, we show that our proposed algorithm converges at a linear rate to the neighborhood of the optimal solution.
\item[3)] We provide extensive numerical experiments on both  i.i.d. and non-i.i.d. data distributions. The experimental results show the robustness of our proposed approach to different types of Byzantine attacks.
\end{itemize}
The remainder of the paper is organized as follows. In the next section, we introduce analog OTA-FL and the transmission model. In Section \ref{Proposed_approach}, our proposed framework is presented. In section \ref{noniid}, our approach is extended to handle non-i.i.d. data distributions among clients. Numerical experiments are provided in Section \ref{numerical_simulation} and concluding remarks are drawn in Section \ref{conclusion}.

%\section{Related Work}

\section{System Model}
\label{system_model}
We consider a FL system where $N$ clients, each with a local dataset $D_n = \{(\bx_i\in \mathbb{R}^d ,y_i \in\mathbb{R})\}_{i=1}^{m_n}$, communicate with a CS to collaboratively train a global model. The output of the FL process is the optimal parameter vector  $\bw^\star\in\mathbb{R}^p$ that minimizes a global loss function $f(\bw)$ given by
\begin{equation}
f(\bw)=\frac{1}{N}\sum_{n=1}^N \ex_{\boldsymbol\bxi \sim \mathcal{D}_n}f_n(\bw, \bxi),
\end{equation}
where $f_n(\bw,\bxi)$ is the local loss function at client $n$, $\mathcal{D}_n$ is the data distribution of the local dataset $n$. %\begin{equation}
%f_n(\bw)=\frac{1}{m_n}\sum_{j=1}^{m_n} \ell(\bw; \bx_j,y_j).
%\end{equation}
%Usually, the CS communicates with the clients using wireless channels. We consider in this work the case of analog OTA-FL \cite{9042352}, which will be introduced hereafter. 
One of the main challenges of FL is the communication bottleneck, where the parameter updates (or the stochastic gradients) of the participating clients need to be sent to the server at each global training round, and usually over limited bandwidth wireless channels. This has motivated researchers to propose over the air computation as a promising solution to reduce the communication overhead. In the sequel, we introduce analog OTA-FL, where the parameter updates are sent simultaneously over the multiple access channel.
\subsection{Analog over-the-air FL}
By virtue of its communication-efficiency, OTA-FL has attracted increasing interest from many researchers. Different variants of OTA-FL have been proposed considering different design criteria, such as power control, scheduling, beamforming design, learning rate optimization and gradient compression. Our proposed framework can be integrated with any of these variants. But, for sake clarity, we adopt the OTA-FL design proposed in \cite{Sery2020}.

For OTA-FL, at each global training round $t$, the CS sends the model parameter vector, $\bw_t$, to the clients. It is usually assumed that the downlink communication is perfect due to the high power available at the CS. Thus, each client receives the global model without distortions. Then, client $n$  sets its local model as $\bw_{t,0}^n=\bw_t$ and runs its local SGD for $H$ iterations based on its local dataset 
\begin{equation}
\bw_{t,i+1}^n=\bw_{t,i}^n-\eta_t  f_{n,j_{n,i}^t}'(\bw_{t,i}^n), \ \ {\rm for} \ \ i = 0,1,\cdots, H-1,
\label{SGD}
\end{equation}
where $\eta_t$ is the SGD step size at round $t$ and $f_{n,j_{n,i}^t}'(\bw_{t,i}^n)$ denotes the stochastic gradient computed using a sample with index $j_{n,i}^t$ chosen uniformly at random from the local dataset of client $n$. In practice, a minibatch can be used instead of a single sample to compute the stochastic gradient.
The clients then send their model updates,
\begin{equation}
\bm_t^n = \bw_{t,H}^n - \bw_{t},
\end{equation}
simultaneously to the CS via analog OTA. The model updates should be precoded in order to mitigate the effect of channel fading.  Let $ \tilde h_{n,t}= h_{t,n}e^{j \phi_t^n}$ be the block fading channel corresponding to user $n$ at the transmission time of global round $t$, where $h_{t,n}  >0$ and $\phi_t^n=[-\pi,\pi]$ are its module and phase respectively. As in \cite{Kai2020,9076343,9042352,Sery2020,Liu2021}, we assume that perfect channel state information (CSI) is available at the clients and the CS. The imperfect CSI case is left for future investigation. Moreover, since the power budget at the clients is limited, the transmitted signal should satisfy the following average power constraint
\begin{equation}
\mathbb{E} \left[ \| \bx_{t,n} \|^2 \right] \leq P.
\label{power_const}
\end{equation}

In practice, weak channels might cause a high amplification of transmit power, possibly violating the transmission power constraint \eqref{power_const}. To overcome this issue, a threshold $h_{\min}$ can be set and clients with channel fading coefficients less than $h_{\min}$ in magnitude will not transmit in that training round.

We adopt in this paper the precoding scheme proposed in \cite{Sery2020}, where every client $n$ precodes its model update $\bm_t^n$ as
\begin{equation}
 \bx_{t,n}=\begin{cases}\rho_{t}\frac{h_{ \rm min}}{ h_{t,n}}   e^{-j \phi_t^n}   \bm_t^n, \ \ \  {\rm if} \ \  h_{t,n}>h_{\min}\\ 0, \ \ \ \ \ \ \ \ \ \ \ \ \    \ \ \ \ \ \ \ {\rm if} \ \  h_{t,n} \leq h_{\min}\end{cases}
 \label{precoding}
\end{equation}
where $\rho_{t}$ is a factor to satisfy the average power constraint at client $n$, which is set as follows 
\begin{equation}
\rho_{t} =\sqrt{ \frac{P}{ \max_{n} \mathbb{E}  \| \bm_t^n \|^2}}.
\end{equation}
%In practice, $\bx_n$ is sent over a time slot $T$ where the following analog signal is sent
%$$
%x_n(t)=\rho_{k,n}\frac{h_{k,n}^*}{| h_{k,n}|^2}<\bs(t),\bw_{k+H}^n)>
%$$
%$\bs(t)=[s_1(t),\cdots,s_d(t)]^T$ is a set of orthonormal waveforms. Once $\{x_n(t)\}_{n=1}^N$ are computed, the clients send simultaneously their corresponding signals. The CS receives 
%$$
%y(t)=\sum_{n=1}^N  h_{n,t} x_n(t)+z(t)
%$$
%where $z(t)$ stands for the additive noise. Filtering with the corresponding waveforms, 
Note that in practice, the clients do not have access the updates of each  other in order to compute $\rho_t$ at each training round. A possible way to deal with this issue is that  the CS can estimate this parameter offline using a small dataset and then forward it to the clients so they can use it at every training round. Another solution is to obtain an upper bound for $\rho_t$ and use it at every iteration \cite{Sery2020}.
The received signal at the CS is 
\begin{equation}
\by_t=\sum_{n \in \mathcal{K}_t} \rho_t h_{\min}\bm^n_{t} +\tilde\bz_t,
\end{equation}
where $\mathcal{K}_t$ is the set of clients indices with channel fading verifying $h_{n,t}>h_{\min}$ and $\tilde\bz_t \sim \mathcal{N}(\boldsymbol{0},\sigma^2\bI_p)$ stands for additive noise. In order to update the global model, the CS sets
\begin{equation}
\bw_{t+1}=\frac{\by_t}{|\mathcal{K}_t|\rho_t h_{\min}}+\bw_t,
\label{global_update}
\end{equation}
where $|\mathcal{K}_t|$ is the cardinality of the set $\mathcal{K}_t$. The global model update in \eqref{global_update} can be also written as
\begin{equation}
\bw_{t+1}=\frac{1}{|\mathcal{K}_t|}\sum_{n=1}^{|\mathcal{K}_t| }\bw^n_{t,H} + \bz_t,
\end{equation}
where $\bz_t \triangleq  \frac{\tilde\bz_t}{|\mathcal{K}_t|\rho_t h_{\min}} \sim \mathcal{N}(\boldsymbol{0}, \frac{\sigma^2}{ |\mathcal{K}_t|^2 \rho_t^2 h_{\min}^2 } \bI_p)$.
\subsection{Byzantine attacks}

Although FL addresses the issue of sending sensitive data of the clients, it opens the way to possible adversarial attacks since it allows the clients to perform model update. A popular class of adversarial attacks in this context is Byzantine attacks where some clients send falsified parameter updates aiming at affecting the convergence of the training phase or promoting a particular model. Byzantine attacks include also modifications due to possible hardware failures for example. The challenge of designing a Byzantine robust FL process have attracted many researchers  \cite{blanchard2017machine,Xie2018GeneralizedBS,minsker2015geometric,10.1145/3154503,yin2018byzantine,10.1145/3322205.3311083,li2019rsa,chen2018draco,Rajput2019DETOXAR,yin2019defending,9153949}. However, these works are applicable in the case where the model updates of the clients are sent separately which is not the case of OTA-FL. The challenge becomes more significant in the case of OTA-FL where all the model updates are sent simultaneously over the analog wireless channel. This may affect heavily the applicability of OTA-FL in practice. In the next section, we propose a transmission and aggregation framework to deal with Byzantine attacks in the context of OTA-FL.

%We assume that $B<N$ clients are malicious; sending arbitrary or modified parameter vector updates aiming at affecting the convergence of the global model or forcing it to converge to some particular solution. This type of attacks is known as Byzantine attacks. There are many works that proposed solutions to deal with this type of attacks in federated learning \cite{blanchard2017machine,Xie2018GeneralizedBS,minsker2015geometric,10.1145/3154503,yin2018byzantine,10.1145/3322205.3311083,li2019rsa,chen2018draco,Rajput2019DETOXAR,yin2019defending,9153949}. However, these works considered the case of wired FL or the case that each individual update of the client is sent separately to the CS. In this work, we consider the effect of Byzantine attacks in the context of analog OTA-FL and we propose a practical approach to deal with such attacks. 

\section{Byzantine Resilient OTA-FL}
\label{Proposed_approach}
In this section, we first develop our approach under the assumption that the local datasets at all the clients are identically distributed and then analyze its convergence. The extension to the case of non-i.i.d. data is considered in the next section.
\subsection{Algorithm development}
We assume that $B$ clients are Byzantine and the rest $R=N-B$ clients are regular.
In order to reduce the effect of Byzantine attacks, we propose the following approach. At each global training round $t$, the CS divides uniformly at random the $N$ clients into $G=N/m$ groups where each group is composed of $m$ clients. Each group will be allocated a time slot for transmission of their model updates. Precisely, the clients of group $g$ will transmit simultaneously their updates over-the-air. This allows the CS to obtain $G$ model updates. Then, with a robust aggregation technique, the different model updates of the groups will be aggregated to update the global model. We will demonstrate that this approach will be robust to Byzantine attacks.

At global iteration $t$, the global model, $\bw_t$, is forwarded to all the clients. The clients in group $g$ perform $H_g$ steps of SGD using their local datasets as in $\eqref{SGD}$, and compute their model updates
\begin{equation}
\bm_t^n = \bw_{t,H_g}^n - \bw_{t}, \ \ {\rm for} \ \  n \in \mathcal{G}_{g,t},
\end{equation}
where $\mathcal{G}_{g,t}$ is the set of user indices belonging to group $g$ at the global training round $t$. Note that since the clients in different groups are not sending at the same time slot, we can let the number of the SGD steps varying among groups. In other terms, the clients of a group with later transmission time can perform more SGD steps than those in the current transmitting group. However, for simplicity we assume in the sequel that all clients perform the same number of SGD steps $H$ regardless of their transmission time.  
The clients in group $g$ compute their precoded signal as 
\begin{equation}
 \bx_{t,n}=\begin{cases}\rho_{t}\frac{h_{ \rm min}}{ h_{t,n}}   e^{-j \phi_t^n}   \bm_t^n, \ \ \  {\rm if} \ \  h_{t,n}>h_{\min}\\ 0, \ \ \ \ \ \ \ \ \ \ \ \ \  \   \ \ \ \ \ {\rm if} \ \  h_{t,n} \leq h_{\min}\end{cases}
 \label{precoding_11}
 \end{equation}
and send their updates simultaneously during their allocated transmission time slot $T_g$. At the CS, the received signal vector corresponding to group $g$ can be expressed as
\begin{equation}
\by_{t,g}=\sum_{n \in \mathcal{K}_{t,g}} \rho_t h_{\min}\bm^n_{t} +\tilde\bz_{t,g},
\end{equation}
where $\mathcal{K}_{t,g}$ is the set of clients indices of group $g$ with channels such that $h_{n,t}>h_{\min}$ and $\tilde\bz_{t,g} \sim \mathcal{N}(\boldsymbol{0},\sigma^2\bI_p)$ is the additive noise. The CS estimates the model update of the group $g$ as
\begin{equation}
\bu_g ^t= \frac{\by_{t,g}}{\rho_t h_{\min} |\mathcal{K}_{t,g}|}.
\label{group_update}
\end{equation}
After all the group updates are collected, the CS disposes of $G$ vector updates $\bu_1^t,\cdots,\bu_G^t$ and can aggregate these updates to obtain the global model. For instance, one of the most efficient aggregation techniques that can be used is the geometric median \cite{bhagoji2019analyzing}. Other aggregation techniques can be used such as the Krum aggregation rule proposed in \cite{blanchard2017machine}. In this work, we focus on geometric median aggregation. 
The global model is updated as
\begin{equation}
\bw_{t+1} = \gm(\bu_1^t,\cdots,\bu_G^t) + \bw_t,
\label{global_update_}
\end{equation}
where $\gm(.)$ stands for the geometric median aggregation defined as
$$
\gm(\{ \bu_i\}_{i\in \mathcal{K}}) = {\rm arg} \min_\bz \sum_{i\in \mathcal{K}}  \|\bz-\bu_i\|.
$$
\begin{algorithm}[ht]
  \SetAlgoLined
  \caption{ROTAF}
  \BlankLine
\KwIn{Initial model $\bw_0$}
\For{$t = 0 , 1, 2 ,\cdots$}{The CS forwards $\bw_t$ to the clients\;
%divides the clients into $G$ groups uniformly at random and assigns a transmission time for each group
\For{each client $n$}{
$\bm_t^n\gets LocalComp(\bw_t,H, b,\eta,D_n)$
}
\For{$g = 1,2,\cdots,G$}{
	\For{each client $n $ in group $g$ $(n\in \mathcal{G}_{g,t}) $}{
	client $n$ transmit its model update $\bx_{t,n}$ precoded via \eqref{precoding_11} during transmission time slot $T_g$
	}
	The CS receives $\by_g$ of group $g$ and computes $\bu_t^g$ as in \eqref{group_update}
}
The CS aggregates the group updates using \eqref{global_update_} to obtain the new global parameter vector $\bw_{t+1}$
}
\label{sum_alg}
\end{algorithm}
The geometric median  aggregation has been proposed as an efficient solution to Byzantine attacks when the individual updates of the clients are sent separately to the CS \cite{li2019rsa,9153949}. In fact, it approximates well the mean of the honest clients weight updates when $B<N/2$ \cite{9153949}. In our approach, it is used to aggregate the group updates. Thus, the number of Byzantine clients should satisfy $B<G/2$ in order for the geometric median to well approximate the mean of the group updates composed by only regular clients.

The geometric median can be efficiently computed using Weiszfeld \cite{weiszfeld1937point}. To avoid numerical instabilities, a smoothed version of the Weiszfeld algorithm can be used in practice \cite{pillutla2019robust}, which computes a smoothed geometric median defined as 
$$
\gm_{\epsilon}(\{ \bu_i\}_{i\in \mathcal{K}}) = {\rm arg} \min_\bz \sum_{i\in \mathcal{K}}  \|\bz-\bu_i\|_\epsilon,
$$
where
$$
\|\bx\|_\epsilon=\begin{cases}
\frac{1}{2\epsilon}\|\bx\|^2+\frac{\epsilon}{2} \ \ \ {\rm if} \ \ \|\bx\|\leq\epsilon\\ 
\|\bx\| \ \ \ \ \ \ \ \ \ \ \ \   {\rm if} \ \ \|\bx\|>\epsilon,
\end{cases}
$$
where $\epsilon>0$ is a smoothing parameter. For sake of completeness, we present hereafter the smoothed Weiszfeld algorithm.
The steps of the proposed approach are summarized in Algorithm \ref{sum_alg}, where $LocalComp(\bw_t,H,b,\eta,D_n)$ consists of $H$ steps  of batch-SGD using local dataset $D_n$ with learning rate $\eta$ and initial parameter vector $\bw_t$ as described in \eqref{SGD}.

\begin{algorithm}
  \SetAlgoLined
  \caption{Smoothed Weiszfeld algorithm}
  \BlankLine
\KwIn{ Set of vectors $\{\bv_i\}_{i=1}^k$, smoothing parameter $\epsilon$}
\kwInit{$ \bz = \bw_t$}
\Repeat{$\bz$ converges}{\For{$i=1,\cdots,k$}{$ \theta_i= \frac{1}{\max(\epsilon, \|\bz-\bv_i\| )}$}
$\bz =\frac{ \sum_{i=1}^k  \theta_i \bv_i}{\sum_{i=1}^k  \theta_i}$
}

\label{Weiszfeld_algorithm}
\end{algorithm}

\subsection{Convergence Analysis}
\label{conver_analysis}
In this section, the convergence of the proposed framework is studied. As in most of the works that studied the convergence of SGD, the analysis is conducted under the following assumptions.
\begin{assumption}
\begin{itemize}
\item[(i)] Strong convexity: The objective function $f$ is $\mu$-strongly convex, that is, for all $\bx,\by \in \mathbb{R}^p$
$$
f(\bx)\geq f(\by) + \langle f'(\by), \bx-\by\rangle+\frac{\mu}{2}\|\bx-\by\|^2.
$$
for some $\mu>0$.
\item[(ii)] Lipschitz continuity of gradients: The objective function $f$ has $L$-Lipchitz continuous gradients, that is, for all $\bx,\by \in \mathbb{R}^p$
$$
\|f'(\bx)-f'(\by)\| \leq L \|\bx-\by\|.
$$
for some $L>0$.
\item[(iii)] Bounded outer variation: For each honest client $n$, the variations of its aggregated gradients with respect to the over-all gradient is bounded as
$$
\| f'_{n}(\bx)-f'(\bx) \|^2 \leq \delta^2,   \  {\rm for   \ all} \  \bx \in \mathbb{R}^p.
$$
\item[(iv)] Bounded inner variation: for each honest client $n$, the variation of its stochastic gradients with respect to its aggregated gradients is bounded as
$$
\mathbb{E}\|f'_{n,j_{n,i}^t}(\bx)-f_n'(\bx)\|^2\leq \kappa^2,  \  {\rm for   \ all} \  \bx \in \mathbb{R}^p.
$$
\item[(v)] Bounded stochastic gradients: For each honest client $n$, stochastic gradient $f'_{n,j_{n,i}^t}(\bx)$ satisfies
$$
\mathbb{E}\|f'_{n,j_{n,i}^t}(\bx)\|^2\leq K^2, \  {\rm for   \ all} \  \bx \in \mathbb{R}^p,
$$
for some fixed $K^2>0$.
\end{itemize}
\label{assump1}
\end{assumption}
Items (i) and (ii) in Assumption \ref{assump1} are common in convex analysis. Items (iii) and (iv) are needed to bound the inner and outer variations of the stochastic gradients and the gradients of the honest clients, respectively. These assumptions are adopted in most of the existing works considering distributed SGD in presence of Byzantine attacks \cite{9153949,pmlr-v80-tang18a}. The convergence of the proposed approach is presented in the following theorem, proved in the Appendix. For simplicity, we assume in this section that the learning rate is constant and that the clients perform one SGD step at each global iteration, that is, $H=1$ and $\eta_t=\eta$ for all $t$.
\begin{theorem} Under Assumption \ref{assump1}, when the number of Byzantine attackers satisfies $B< \frac{G}{2}$ and the step size $\eta$ verifies $\eta< \min(\frac{\mu}{2L^2} , \frac{2}{\mu})$, then 
$$
\mathbb{E}\|\bw_t-\bw^*\|^2\leq  ( 1-\eta\mu)^{t} B_1+A_1,
$$
where 
\begin{align}
B_1&= \|\bw_0-\bw^*\|^2 -A_1,\\
A_1&= \frac{2}{\mu^2}C_\alpha^2 \left(\delta^2+\kappa^2+\frac{p\sigma^2}{mPh_{min}^2} K^2\right),\label{err1}
\end{align}
with $C_\alpha=\frac{2-2\alpha}{1-2\alpha}$ and $\alpha= \frac{B}{G}$.
\label{thm1}
\end{theorem}

Theorem \ref{thm1} states that the proposed approach converges at a linear rate to a neighborhood of $\bw^*$. The asymptotic learning error, $A_1$, depends on the number of Byzantine attackers through $C_\alpha$ and on the noise variance. $C_\alpha$ increases with the number of Byzantine attackers, which yields a higher asymptotic error. Moreover, the error is composed of three terms which are proportional to the outer variations of the gradients, the inner variations of the stochastic gradients, and the noise variance, respectively.

\section{Byzantine resilience for non-i.i.d. data}
\label{noniid}
In the previous section, we have focused on the case of identically distributed local datasets. However, this is not the case in most real applications of federated learning \cite{smith2017federated}. In this section, we modify our proposed approach to handle the case of non-i.i.d. data. 

\subsection{Resampling before aggregation}

The main concern of non-i.i.d. datasets is that the model updates (or the stochastic gradients) are not identically distributed and may have a large variance. Unfortunately,  large variance can affect heavily the performance of most of the existing robust aggregation techniques including the geometric median, Krum, and coordinate wise median \cite{karimireddy2021byzantinerobust} . Recently, a simple solution has been proposed in \cite{karimireddy2021byzantinerobust} to robustify the aggregation techniques in the case of non-i.i.d. data. The resampling process consists of multiple rounds where $s$ vectors of the model updates are sampled uniformly at random at each round, with the constraint that each vector is sampled at most $s$ times. The average of these samples is then computed to generate a new message. The new messages are then fed to the aggregation technique. The steps of the resampling process are presented in Algorithm \ref{sam_alg}.

%{\color{blue}Add a paragraph where the resampling process will be added in the proposed approach}

Since our proposed framework is based on geometric median, its performance can be degraded heavily in the case of non-i.i.d. data. To address this issue, we propose to apply the resampling process on the group updates before computing their geometric median. To motivate the importance of reducing the variance of the model updates before aggregation, we review the following lemma from \cite{9153949} that characterizes the error of the geometric median compared to the true mean.

\begin{lemma}\cite[Lemma 1]{9153949} Let $\mathcal V$ be a subset of random vectors distributed in
a normed vector space with cardinality $N$. Let $\mathcal{B} \subset \mathcal{V}$ be a subset of $\mathcal V$ with cardinality $B$ such that $B< \frac{N}{2}$ and let $R=N-B$. Then, it holds that
$$
\mathbb{E}\|\underset{\bv\in \mathcal{V}}\gm (\bv) -\bar\bv \|^2 \leq \frac{ C_\alpha^2}{R}{\sum_{\bv \notin \mathcal{B}}\|\bv -\mathbb{E}\bv\|^2}+\frac{ C_\alpha^2}{R} {\sum_{\bv \notin \mathcal{B}}\|\ex\bv -\bar\bv\|^2},
$$
where $\bar\bv = \frac{1}{R}\sum_{\bv \notin \mathcal{B}} \ex\bv$, $C_\alpha=\frac{2-2\alpha}{1-2\alpha}$ and $\alpha= \frac{B }{N}$.\label{gm_Byzantine}\end{lemma}

\begin{algorithm}
  \SetAlgoLined
  \caption{$s$-resmapling process}
  \BlankLine
\KwIn{ Set of vectors $\mathcal{S} = \{\bv_i\}_{i=1}^R$ of size $R$, resampling rate $s$} 
\KwOut{ A new set of vectors $\tilde{ \mathcal{S}} = \{ \tilde \bv_i \}_{i=1}^R$ }
\For{$i = 1 ,  2 ,\cdots, R$}{Choose uniformly at random $s$ vectors from $\mathcal{S}$ such that any vector is chosen at most $s$ times \;
Compute a new $\tilde \bv_i$ as the average of the chosen vectors $\tilde\bv_i =\frac{1}{s} \sum_{j\in \mathcal{S}_i}\bv_j$
}
\label{sam_alg}
\end{algorithm}

The bound on the error of the geometric median reported in Lemma \ref{gm_Byzantine} is composed of two terms. The first one is related to the inner variations of the vector updates of the regular updates, that is, the stochastic gradient computed at a certain regular client compared to its true gradient. The inner variations can be reduced by increasing the size of the minibatch at every iteration for instance. The second term is related to the outer variations, where the effect the difference between the distribution of the local datasets is highlighted. The resampling process helps reducing the outer variations in the case of non-i.i.d. data. To motivate further the importance of the resampling step, we compare in Table \ref{table:resamp} the performance of our proposed approach, ROTAF, with and without resampling for different number of Byzantine attacks. From the table, without resampling, the performance of the proposed  is heavily degraded for non i.i.d data even in the case where no Byzantine attacks are applied since the geometric median is sensitive to high variance of the input vectors. Moreover, we remark that with resampling the performance is enhanced and approaches that obtained in the case of i.i.d. data distributions.

\begin{table}[ht]
\centering
\caption{Test accuracy of ROTAF for different values of the resampling rate $s$ on non-i.i.d. MNIST dataset. The simulation settings are detailed in section \ref{numerical_simulation}. The test accuracy after 500 global iterations is reported.} 
\subtable[Logistic regression]{
\centering
\begin{tabular}{c c c c c} % centered columns (4 columns)
\hline\hline %inserts double horizontal lines
Number of attacks & non-i.i.d. $s  = 1$ & non-i.i.d. $s =2$  & non-i.i.d. $s=3$  &  i.i.d. $s=1$ \\ [0.5ex] % inserts table
%heading
\hline % inserts single horizontal line
No Byzantine attacks & 0.6842 & 0.8690  & 0.8971  & 0.9151\\ % inserting body of the table
2 Byzantine attackers & 0.6842 & 0.8817 & 0.9090 & 0.9139\\
5  Byzantine attackers & 0.6996 & 0.8829 &0.9102  &0.9112\\  [1ex] % [1ex] adds vertical space
\hline %inserts single line
\end{tabular}
}
\hfill
 \subtable[CNN model]{
\centering
\begin{tabular}{c c c c c} % centered columns (4 columns)
\hline\hline %inserts double horizontal lines
Number of attacks & non-i.i.d. $s  = 1$ & non-i.i.d. $s =2$  & non-i.i.d. $s=3$  &  i.i.d. $s=1$ \\ [0.5ex] % inserts table
%heading
\hline % inserts single horizontal line
No Byzantine attacks & 0.6557 & 0.9034  & 0.9379 & 0.9529 \\ % inserting body of the table
2 Byzantine attackers & 0.6799 & 0.9260 & 0.9533 & 0.9528 \\
5  Byzantine attackers & 0.6814 & 0.9280 & 0.9553  & 0.9526 \\  [1ex] % [1ex] adds vertical space
\hline %inserts single line
\end{tabular}
}
\label{table:resamp} % is used to refer this table in the text
\end{table}
\subsection{Convergence Analysis}
The convergence of our proposed approach for non-i.i.d. data among clients is quite different from that of i.i.d. data due to the resampling step before the geometric median aggregation. In the following theorem, the convergence of our proposed framework with resampling is studied under the assumptions stated in Section \ref{conver_analysis}. The proof is given in the Appendix.
\begin{theorem} Under Assumption \ref{assump1}, when the number of Byzantine attackers satisfies $B < \frac{G}{2s}$ and the step size $\eta$ verifies $\eta< \min(\frac{\mu}{2L^2} , \frac{2}{\mu})$, then 
$$
\mathbb{E}\|\bw_t-\bw^*\|^2\leq  ( 1-\eta\mu)^{t} B_2+A_2,
$$
where 
\begin{align}
B_2&= \|\bw_0-\bw^*\|^2 -A_2,\\
A_2 &= \frac{2}{\mu^2}C_{s\alpha}^2  \left(d+ \frac{1-d}{G-B}\right)\left(\delta^2+\kappa^2+\frac{p\sigma^2}{mPh_{min}^2} K^2\right),
\label{err2}
\end{align}
with $C_{s\alpha}=\frac{2-2s\alpha}{1-2s\alpha}$, $\alpha= \frac{B}{G}$ and $d = \frac{G-1}{Gs-1}$.
\label{thm3}
\end{theorem}

Theorem \ref{thm3} establishes the convergence of the proposed framework for non-i.i.d. data to a neighborhood of the optimal solution with an asymptotic learning error, $A_2$, that depends on the number of Byzantine attackers, the resampling rate and the noise variance. Note that the resampling step comes at the cost of tolerating lower number of attackers $B<\frac{G}{2s}$. This is due to the fact that more vectors will be contaminated by malicious updates as the resampling rate $s$ increases. This is not a major issue since low value of $s$ is needed as we will see later in the numerical experiments. Specifically, $s=2$ or $3$ would be sufficient to robustify the geometric median aggregation against high variance of the model updates in non-i.i.d. data settings.

By comparing the asymptotic errors in \eqref{err1} and \eqref{err2}, the difference is in the coefficients $C_{\alpha}^2$ and $C_{s\alpha}^2  \left(d+ \frac{1-d}{G-B}\right)$, respectively. For $s=1$, the asymptotic errors are equal, while for $s>1$ the coefficient $C_{s\alpha}^2  \left(d+ \frac{1-d}{G-B}\right)\approx C_{s\alpha}^2 d$ is smaller than $C_{\alpha}^2$ when $\alpha$ is sufficiently small. This implies that the asymptotic error is reduced after resampling. However, this comes at the cost of tolerating a smaller number of Byzantine attacks.

\section{Numerical Results}
\label{numerical_simulation}
In this section, the performance of the proposed approach is studied using real datasets, and compared with the existing OTA-FL design.
We consider two popular image-classification datasets, MNIST and CIFAR10.\\
{\bf MNIST: } the dataset is composed of 28x28 images of handwritten digits corresponding to 10 classes. The dataset contains 60,000 training samples and 10,000 testing samples. For the MNIST dataset, we used the multi-class logistic regression model and a CNN model composed of two 5x5 convolutional layers (the first with 32 channels, the second with 64 channels, each followed by 2x2 max pooling), a fully connected layer with 128 neurons and Relu activation, and a Softmax output layer.\\
{\bf CIFAR10:} the dataset is composed of 32x32x3 colour images corresponding to 10 classes. The dataset contains 50,000 training images and 10,000 test images. For this dataset, we adopt a CNN model composed three 3x3 convolutional layers (the first with 32 channels, the second with 64 channels, the third with 128 channels, each followed by 2x2 max pooling), a two fully connected layers with 512 neurons and 128 neurons, respectively, and Relu activation, and a Softmax output layer.

\begin{figure}
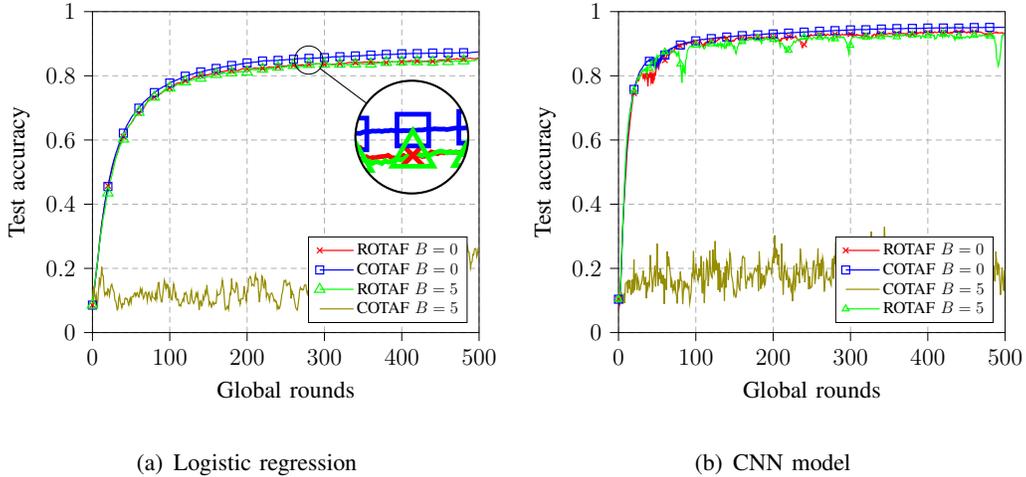

\begin{center}
\subfigure[Logistic regression]{
}

\end{center}
\caption{Test accuracy vs. global training rounds with $B$ Byzantine clients applying Gaussian attacks. MNIST i.i.d. data}
\label{gaussian_attacks}
\end{figure}

We explain hereafter how the training sets are divided among the clients for i.i.d. data and non-i.i.d. data settings:\\
{\bf i.i.d. data:} the whole training dataset is split uniformly at random among clients with the same number of samples per client.\\
{\bf non-i.i.d. data:} First, the classes are sampled with exponentially decreasing portions, that is, $\gamma^i$ portion of samples of class $i$ is taken for certain $\gamma\in (0,1]$. Note that all the classes have the same amount of samples when $\gamma = 1$. The same procedure is applied to the test dataset. Then, the dataset is sorted by its label and equally distributed between clients. This results in clients having samples from certain classes only.

At every global training round, each client performs one step of local batch-SGD, where a minibatch of size $b=50$ is used at each step. The leaning rate $\eta_t$ is fixed to $0.01$ in all local SGD steps. In all experiments, the number of groups is fixed to $G=20$ while the noise variance and the power constraint are set such that $\frac{P}{\sigma^2}=20dB$, unless differently stated. The smoothing parameter of the Weiszfeld algorithm used is $\epsilon=10^{-4}$. In all experiments, the transmission threshold $h_{\min}$ and the scaling factor $\rho_t$ are fixed to 0.1 and 10 respectively.

We consider different types of attack to study the robustness of the proposed approach:\\
{\bf Gaussian attacks}: each Byzantine client sends a Gaussian vector with entries having mean $0$ and variance $30$ instead of its actual model update. \\
{\bf Class-flip attacks}: each Byzantine client changes the labels of its local datasets as $y=9-y$.\\
{\bf Mimic attacks}: all Byzantine clients choose a regular client and mimic its local update. This results in a consistent bias towards emphasizing the chosen worker and thus underrepresenting other clients. Such attack is applied in the case of non-i.i.d. data.

In \figref{gaussian_attacks}, we compare our proposed approach and the simple averaging COTAF \cite{Sery2020} described in Section \ref{system_model}. The MNIST dataset is used with i.i.d. local data at the clients. As expected, the performance of COTAF is heavily affected by Byzantine attacks even for small number of attackers. On the other hand, the proposed approach guarantees fast convergence and provides almost the same performance as the case without attacks. Moreover, the performance of both approaches is the same in the case of no attacks $(B=0)$.

\begin{figure}[ht]
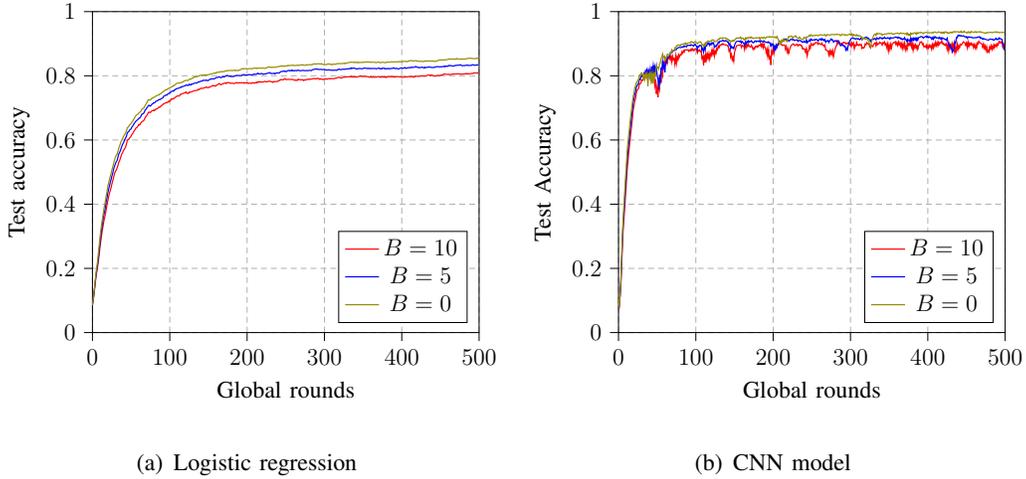

\begin{center}
\subfigure[Logistic regression]{
}
\end{center}
\caption{Test loss and test accuracy vs. the number of iterations for different numbers of Byzantine clients applying class-flip attacks.}
\label{class_flip_attacks}
\end{figure}

In the second experiment, we consider class flip attacks where the Byzantine clients change the labels of their local datasets as $y=9-y$. \figref{class_flip_attacks} demonstrates the effect of increasing the number of Byzantine workers. From this figure, the proposed approach is robust to Byzantine attacks even though the test accuracy relatively decreases with the number of the Byzantine clients. This is expected since the training, in presence of malicious clients, is done over a smaller sample size as only the honest clients contribute in the learning process. This was also predicted by the convergence analysis conducted in section \ref{conver_analysis}, where it has been shown that the asymptotic learning error increases with the number of Byzantine attacks. 

\begin{figure}
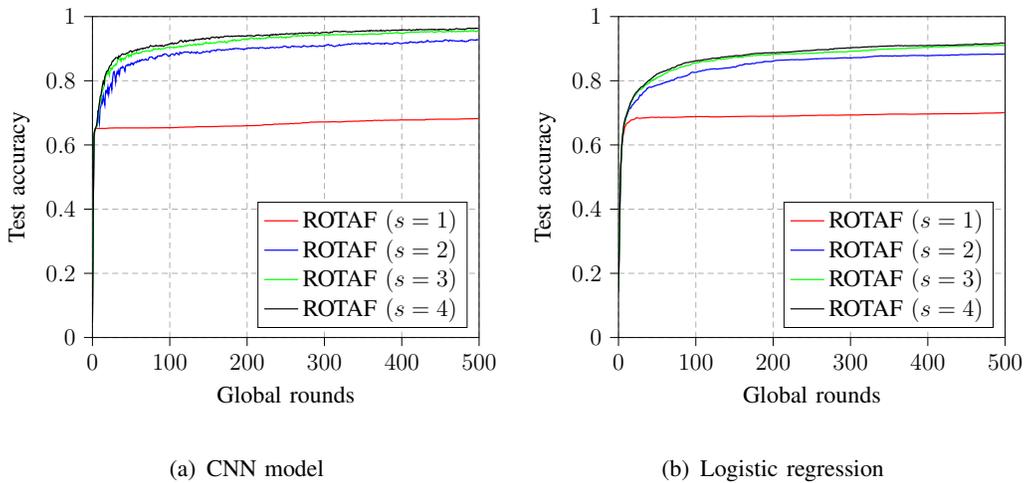

\begin{center}
\subfigure[CNN model]{
%
}
\end{center}
\caption{Test accuracy vs. global rounds for MNIST non-i.i.d. dataset when $\gamma = 0.6$, $P=1$, and $\sigma^2=0$. The performance of the proposed approach with $(s>1)$ and  without $(s=1)$ resampling is reported.}
\label{sample_dup_attack_noniid}
\end{figure}

In the third experiment, we study the effect of non i.i.d data distributions among clients. Particularly, we plot in \figref{sample_dup_attack_noniid} the performance of the proposed approach for MNIST non-i.i.d. dataset in presence of $B=5$ attackers applying mimic attacks. As seen, the performance of ROTAF is significantly enhanced using resampling. Moreover, we note that $s=3$ is sufficient without need for higher values of the resampling rate.

\begin{figure}
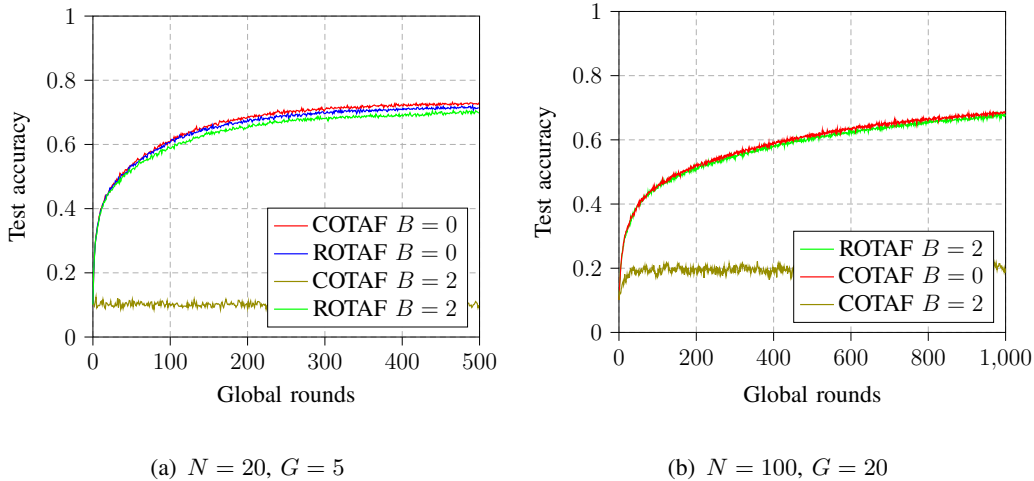

\begin{center}
\subfigure[$N=20$, $G=5$]{
%
}
\end{center}
\caption{Test accuracy vs. global rounds on CIFAR10 i.i.d. dataset when $B$ Byzantine clients applying Gaussian attacks.}
\label{gaussian_attacks_cifar}
\end{figure}

In \figref{gaussian_attacks_cifar}, we compare the proposed framework to COTAF using CIFAR10 i.i.d. dataset. As expected, the performance of COTAF is heavily affected by Byzantine attacks while the performance of ROTAF is similar to the attack-free case.

\section{Conclusion}
\label{conclusion}
In this paper, we have proposed a novel framework to account for Byzantine attacks in OTA-FL. By dividing the distributed clients into groups, the parameter server is able to reduce the effect of Byzantine attacks by aggregating the group parameter updates. The convergence of the proposed algorithm has been studied analytically. We have extended our approach to handling the case of non-i.i.d. data. The simulation results show the robustness of the proposed method to different Byzantine attacks and its convergence for both i.i.d. and non-i.i.d. data. This work can be extended by studying other robust aggregation techniques rather than the geometric median. Considering imperfect CSI is also a possible future direction.

\section*{Appendix}
\subsection{Proof of Theorem \ref{thm1}}
To simplify notations, we assume further that all the clients in each group are transmitting at each global training round, that is, $|\mathcal{K}_{t,g}|= |\mathcal{G}_{t,g}|=m$. We think that the extension to the general case is easy and the final convergence results will be the same. We first sate the following lemma which will be used later in the proof.
\begin{lemma}\cite[Lemma 2]{9153949} Let $\mathcal V$ be a subset of random vectors distributed in
a normed vector space. If $\mathcal{V}' \subset \mathcal{V}$ such that $|\mathcal{V}'| < \frac{|\mathcal{V}|}{2}$,
then it holds that
$$
\mathbb{E}\|\underset{\bv\in \mathcal{V}}\gm (\bv) \|^2 \leq C_\alpha^2 \frac{\sum_{\bv \notin \mathcal{V}'}\mathbb{E}\|\bv\|^2}{|\mathcal{V}| - |\mathcal{V}'| },
$$
where $C_\alpha=\frac{2-2\alpha}{1-2\alpha}$ and $\alpha= \frac{|\mathcal{V}'| }{|\mathcal{V}| }$.\label{ineq_gm_Byzantine}\end{lemma}
Define $\delta_t \triangleq \mathbb{E}\|\bw_t-\bw^*\|^2$, where the expectation is taken over the stochastic gradients and the channel noise. To prove Theorem \ref{thm1}, we start by finding an upper bound for $\delta_{t+1}$,
\begin{align*}
\delta_{t+1} & = \mathbb{E}\|\bw_{t+1}-\bw^*\|^2\\&= \mathbb{E}\|\bw_t-\eta f'(\bw_t)-\bw^*+\bw_{t+1}-\bw_t+\eta f'(\bw_t)\|^2\\
&\leq \frac{1}{1-\gamma}\mathbb{E}\|\bw_t-\eta f'(\bw_t)-\bw^*\|^2 +\frac{1}{\gamma}   \mathbb{E}\|\bw_{t+1}-\bw_t+\eta f'(\bw_t)\|^2,
\end{align*}
for any $0<\gamma<1$, where we have used the fact that $\|\bx+\by\|^2\leq \frac{1}{1-\gamma}\|\bx\|^2+\frac{1}{\gamma}\|\by\|^2$ for any $0<\gamma<1$. Since $f'(\bw^*)=0$, we can write
\begin{align*}
& \|\bw_t-\eta f'(\bw_t)-\bw^*\|^2\\& =\|\bw_t-\eta ( f'(\bw_t)-f'(\bw^*) )-\bw^*\|^2\\ &= 
\|\bw_t-\bw^*\|^2 -2 \eta \langle f'(\bw_t)-f'(\bw^*), \bw_t -\bw^*\rangle + \eta^2 \|f'(\bw_t)-f'(\bw^*)\|^2
\\ & \overset{(a)}{\leq}  \|\bw_t-\bw^*\|^2 -2 \eta \mu  \|\bw_t-\bw^*\|^2 + \eta^2 L^2 \|\bw_t-\bw^*\|^2
\end{align*}  
where $(a)$ follows from items $(i)$ and $(ii)$ of Assumption \ref{assump1}. Thus,
\begin{align*}
&\delta_{t+1} \leq \frac{1-2\eta\mu +\eta^2L^2}{1-\gamma}\delta_t + \frac{1}{\gamma}   \mathbb{E}\|\bw_{t+1}-\bw_t+\eta f'(\bw_t)\|^2.
\end{align*}  
For  $\eta<\frac{2}{\mu}$, we can take $\gamma = \frac{\eta\mu}{2}$. Assuming further that $\eta \leq \frac{\mu}{2L^2}$, it holds that
$
\frac{1-2\eta\mu +\eta^2L^2}{1-\gamma} \leq 1-\eta\mu
$. Hence, for $\eta <\min(\frac{2}{\mu}, \frac{\mu}{2L^2})$,
\begin{align}
\delta_{t+1}  &\leq ( 1-\eta\mu) \delta_t   +\frac{2}{\eta \mu}   \mathbb{E}\|\bw_{t+1}-\bw_t+\eta f'(\bw_t)\|^2.
\label{eq:102}
\end{align}
We treat now the second term of the right-hand side of \eqref{eq:102}. From the update rule \eqref{global_update_}, it follows that
$
\bw_{t+1} -\bw_{t} = \gm(\{\bu_g^t\}_{g=1}^G)  
$
 where $\bu_g^t$ is given by
 \begin{align*}
 \bu_g ^t&=  -\frac{1}{m} \sum_{n\in \mathcal{G}_{t,g}} \eta f_{n,i_{n}^t}'(\bw_{t}) + \bz_{t,g},
\end{align*}
 Define $\mathcal{B}_t$ as the set of groups containing at least one Byzantine attacker at global iteration $t$ and $\mathcal{R}_t$ as the set of groups without any Byzantine attackers. 
Applying Lemma \ref{ineq_gm_Byzantine} yields
\begin{align*}
& \mathbb{E}\|\bw_{t+1}-\bw_t+\eta f'(\bw_t)\|^2  =  \mathbb{E}\|\gm(\{\bu_g^t\}_{g=1}^G)+\eta f'(\bw_t)\|^2
 =  \mathbb{E}\|\gm(\{\bu_g^t+\eta f'(\bw_t)\}_{g=1}^G)\|^2\\
 & \leq \frac{C_\alpha^2 }{|\mathcal R_t|}{\sum_{g\in \mathcal R_t}\mathbb{E}\left\| \bu_{g}^t + \eta f'(\bw_t) \right\|^2},
%\\& = \frac{C_\alpha^2 }{|\mathcal R_t|}{\sum_{g\in \mathcal R_t}\mathbb{E}\left\| \bu_{g}^t - \ex\bu_g^t \right\|^2} + \frac{C_\alpha^2 }{|\mathcal R_t|} {\sum_{g \in \mathcal R_t}\left\| \ex\bu_{g}^t + \eta f'(\bw_t) \right\|^2}.
 \end{align*}
%where the last equality is obtained by noting that the cross terms vanish. 
Replacing $\bu_{g}^t$ by its expression, it follows that 
{\small
\begin{align}
& \mathbb{E}\|\bw_{t+1}-\bw_t+\eta f'(\bw_t)\|^2 \nonumber \\&\leq   \frac{C_\alpha^2 }{|\mathcal R_t|} {\sum_{g \in \mathcal R_t}\mathbb{E} \left\|-\frac{\eta}{m} \sum_{n \in  \mathcal G_{t,g}}  f'_{n,i_{n}^t}(\bw_{t}) + \bz_{t,g} +\eta f'(\bw_t) \right\|^2} \nonumber \\
 &= \frac{C_\alpha^2 }{m} {\sum_{g \in \mathcal R_t}\mathbb{E} \left\|-\frac{\eta}{m} \sum_{n \in  \mathcal G_{t,g}} ( f'_{n,i_{n}^t}(\bw_{t}) -f'_{n}(\bw_{t})) + \bz_{t,g}  \right\|^2}+ \frac{C_\alpha^2 }{|\mathcal R_t|} {\sum_{g \in \mathcal R_t} \left\|-\frac{\eta}{m} \sum_{n \in  \mathcal G_{t,g}} f'_{n}(\bw_{t})+\eta f'(\bw_t)\right\|^2} 
%\nonumber\\& \leq 4 C_\alpha^2 \eta^2 \frac{\sum_{g \notin \mathcal B_t} \frac{1}{|\mathcal K_{t,g}|} \sum_{n \in  \mathcal K_{t,g}} \mathbb{E}\left\|f'_{n,i_{n}^t}(\bw_{t}^n)- f'_n(\bw_t) \right\|^2}{G-|\mathcal B_t|}
% \nonumber\\& + 4 C_\alpha^2 \eta^2 \frac{\sum_{g \notin \mathcal B_t} \frac{1}{|\mathcal K_{t,g}|} \sum_{n \in  \mathcal K_{t,g}}\mathbb{E}\left\| f'_{n}(\bw_t) - f'(\bw_t) \right\|^2}{|\mathcal R_t|}
%+2 C_\alpha^2\frac{\sum_{g \notin \mathcal B_t}\mathbb{E}\|\bz_{t,g}\|^2}{G-|\mathcal B_t|},
\label{main}
\end{align}}
where the last equality is obtained by noting that the cross terms vanish. First, we note that
\begin{align}
&\mathbb{E} \left\|-\frac{\eta}{m} \sum_{n \in  \mathcal G_{t,g}} ( f'_{n,i_{n}^t}(\bw_{t}) -f'_{n}(\bw_{t})) + \bz_{t,g}  \right\|^2 \\&= \mathbb{E} \left\|-\frac{\eta}{m} \sum_{n \in  \mathcal G_{t,g}} ( f'_{n,i_{n}^t}(\bw_{t}) -f'_{n}(\bw_{t}))  \right\|^2 +\mathbb{E} \left\| \bz_{t,g}  \right\|^2 \label{eq1118}
\\& \leq  \frac{\eta^2}{m}\sum_{n \in  \mathcal G_{t,g}} \mathbb{E} \left\| ( f'_{n,i_{n}^t}(\bw_{t}) -f'_{n}(\bw_{t}))  \right\|^2 +\mathbb{E} \left\| \bz_{t,g}  \right\|^2 \label{eq1119}
\end{align}
where the equality in \eqref{eq1118} follows by noting that the cross terms vanish due to the independence of the noise and the stochastic gradients while the last result follows from the fact that $\left\|\sum_{i=1}^k \bv_i \right\|^2\leq k \sum_{i=1}^k\|\bv_i\|^2$ for any sequence of vectors $\{\bv_i\}_{i=1}^k$. Similarly, we have 
\begin{align}
\left\|-\frac{\eta}{m} \sum_{n \in  \mathcal G_{t,g}} f'_{n}(\bw_{t})+\eta f'(\bw_t)\right\|^2 \leq \frac{\eta^2}{m} \sum_{n \in  \mathcal G_{t,g}} \| f_n'(\bw_t) -f'(\bw_t)\|^2
\end{align}
Thus, \eqref{main} can be written as
\begin{align}
 \mathbb{E}\|\bw_{t+1}-\bw_t+\eta f'(\bw_t)\|^2 \nonumber \leq  & \frac{C_\alpha^2 }{|\mathcal R_t|} {\sum_{g \in \mathcal R_t} \left[ \frac{\eta^2}{m}\sum_{n \in  \mathcal G_{t,g}} \mathbb{E} \left\| ( f'_{n,i_{n}^t}(\bw_{t}) -f'_{n}(\bw_{t}))  \right\|^2 +\mathbb{E} \left\| \bz_{t,g}  \right\|^2 \right] }\nonumber\\&+ \frac{C_\alpha^2 }{|\mathcal R_t|} {\sum_{g \in \mathcal R_t} \frac{\eta^2}{m} \sum_{n \in  \mathcal G_{t,g}} \| f_n'(\bw_t) -f'(\bw_t)\|^2
} 
 \end{align}
 We deal first with the term $\mathbb{E}\|\bz_{t,g}\|^2$,
\begin{align}
&\mathbb{E}\|\bz_{t,g}\|^2= \frac{p\sigma^2}{h_{min}^2 m^2\rho_t^2} =  \frac{p\sigma^2}{h_{min}^2 m^2} \frac{\max_n\mathbb{E}\|\bm_n^t\|^2}{P}\nonumber
\end{align}
Using item (v)  in Assumption \ref{assump1}, we get
\begin{align}
\mathbb{E}\|\bm_n^t\|^2 = \mathbb{E}\left\|  \eta f'_{n,i_{n}^t} (\bw^n_{t})  \right\|^2 \leq \eta^2K^2
\end{align}
Thus,
\begin{align}
\mathbb{E}\|\bz_{t,g}\|^2\leq   \frac{p\sigma^2}{m^2P h_{min}^2} \eta^2K^2.
\label{noise_bound}
\end{align}
%We consider now the term $\mathbb{E}\left\|\sum_{i=0}^{H-1} f'_{n,j_{n,i}^t}(\bw_{t,i}^n)- f'_n(\bw_t) \right\|^2$. Noting that $\bw_{t,0}^n = \bw_t$, we have
%{\small
%\begin{align}
%&\mathbb{E}\left\|\sum_{i=0}^{H-1} f'_{n,j_{n,i}^t}(\bw_{t,i}^n)- f'_n(\bw_t) \right\|^2= \mathbb{E}\left\|\sum_{i=1}^{H-1} f'_{n,j_{n,i}^t}(\bw_{t,i}^n)+ f'_{n,j_{n,0}^t}(\bw_{t})- f'_n(\bw_t) \right\|^2\nonumber\\&=  \mathbb{E}\left\|\sum_{i=1}^{H-1} f'_{n,j_{n,i}^t}(\bw_{t,i}^n) \right\|^2 +   \mathbb{E}\left\|  f'_{n,j_{n,0}^t}(\bw_{t}) - f'_n(\bw_t) \right\|^2+ 2  \mathbb{E} \left\langle \sum_{i=1}^{H-1} f'_{n,j_{n,i}^t}(\bw_{t,i}^n),f'_{n,j_{n,0}^t}(\bw_{t}) - f'_n(\bw_t)\right\rangle\nonumber 
%\\ &\leq\mathbb{E}\left\|\sum_{i=1}^{H-1} f'_{n,j_{n,i}^t}(\bw_{t,i}^n) \right\|^2+   \mathbb{E}\left\|  f'_{n,j_{n,0}^t}(\bw_{t}) - f'_n(\bw_t) \right\|^2\\&+ 2 \sqrt{\mathbb{E}\left\|\sum_{i=1}^{H-1} f'_{n,j_{n,i}^t}(\bw_{t,i}^n) \right\|^2} \sqrt{\mathbb{E}\left\|f'_{n,j_{n,0}^t}(\bw_{t}) - f'_n(\bw_t) \right\|^2}\nonumber
%\\&\leq (H-1)^2K^2+\kappa^2+2(H-1)K\kappa = (\kappa+(H-1)K)^2
%\label{second}
%\end{align}}
From Assumption \ref{assump1}, we have 
\begin{align}
 \mathbb{E}\left\|f'_{n,i_{n}^t}(\bw_{t}^n)- f'_n(\bw_t) \right\|^2\leq \kappa^2.
 \label{second}
 \end{align}
Moreover, the term $\left\| f'_{n}(\bw_t) - f'(\bw_t) \right\|^2$ verifies by Assumption \ref{assump1},
\begin{align}
\left\| f'_{n}(\bw_t) - f'(\bw_t) \right\|^2 \leq \delta^2.
\label{third}
\end{align}
Combining \eqref{main} with \eqref{noise_bound} , \eqref{second} and \eqref{third}, it holds that
\begin{align}
 \mathbb{E}\|\bw_{t+1}-\bw_t+\eta f'(\bw_t)\|^2 \!\leq C_\alpha^2 \eta^2\! \left(\delta^2+\kappa^2+\frac{p\sigma^2}{mPh_{min}^2} K^2\right)\!.
 \label{eq202}
\end{align}
Combining \eqref{eq:102} and \eqref{eq202} yields for $\eta< \min(\frac{\mu}{2L^2} , \frac{2}{\mu})$
$$
\delta_{t+1}\leq  ( 1-\eta\mu)   \delta_t +\eta \mu A_1,
$$
where
$
A_1  \triangleq \frac{2}{\mu^2}C_\alpha^2 \left( \delta^2+\kappa^2+\frac{p\sigma^2}{mPh_{min}^2} K^2\right)
$. Thus,
\begin{align}
\delta_{t+1} & \leq ( 1-\eta\mu)^{t+1} \delta_0 + \eta \mu A_1\sum_{i=0}^{t}( 1-\eta\mu)^{i}\\
& =  ( 1-\eta\mu)^{t+1} ( \delta_0 - A_1)+A_1,
\end{align}
which completes the proof.

\subsection{Proof of Theorem \ref{thm3}}
We state first the following preliminary results that will be used later in the proof.
\begin{lemma}\cite[Proposition 1]{karimireddy2021byzantinerobust} Let $\{\bv_k, k \in \mathcal{G}\}$ be a subset of vectors with cardinality $G=|\mathcal{G}|$ and  $\{\tilde\bv_k, k\in \mathcal{G}\}$ a new set generated using the resampling method with s-replacement. Assume that $R=G-B$ vectors of the set $\{\bv_k, k \in \mathcal{G}\}$ are uncontaminated (not affected by malicious updates) and $\mathcal{R}$ denoting the set of indices of these vectors. When $B<\frac{G}{2s}$, there exist a set $\mathcal{R}' \subseteq \mathcal{G}$ with at least $G-sB$ elements, such that for any $k' \in  \mathcal{R}'$
\begin{align}
 \mathbb{E} \tilde\bv_{k'} = \frac{1}{R}\sum_{k\in \mathcal{R}}\bv_k,\label{res1}
\end{align}
\begin{align}
 \mathbb{E} \| \tilde\bv_{k'} - \mathbb{E}\tilde\bv_{k'}  \|^2 = \frac{d}{R}\sum_{k\in \mathcal{R}} \left\|\bv_k - \frac{1}{R}\sum_{k\in \mathcal{R}} \bv_k\right\|^2,\label{res2}
\end{align}
 where $d:= \frac{G-1}{sG-1}$ and the expectation is taken over the resampling process.
 \label{lem3}
\end{lemma}

\begin{lemma}
Let $\{\bv_k, k \in \mathcal{G}\}$ be a subset of random vectors in a normed vector space and a subset from it composed of independent vectors $\{\bv_k, k \in \mathcal{R}\}$ are independent, where $\mathcal{R}$ is the indices of the vectors of this subset as defined in Lemma \ref{lem3}. Let $\{\tilde\bv_k, k\in \mathcal{G}\}$ be a new set generated from $\{\bv_k, k \in \mathcal{G}\}$ using the resampling method with s-replacement. When $B<\frac{G}{2s}$, there exist a set $\mathcal{R}' \subseteq \mathcal{G}$ with at least $G-sB$ elements, such that 
{
\begin{align}
\frac{1}{R'}\sum_{k'\in\mathcal{R}'}\mathbb{E}\left\|\tilde\bv_{k'}-\bar\bv \right\|^2=&\left(d-\frac{1-d}{R}\right)\frac{1}{R}\sum_{k\in\mathcal{R}}\mathbb{E}\left\|\bv_k-\mathbb E \bv_k\right\|^2+\frac{d}{R}\sum_{k\in\mathcal{R}}\left\| \mathbb E \bv_k -\bar\bv\right\|^2, \label{eqlem45}
\end{align}}
where $\bar\bv=\frac{1}{R}\sum_{n\in \mathcal{R}}\mathbb E \bv_n$, $d= \frac{G-1}{sG-1}$ and the expectation is taken over the resampling process and the randomness of vectors $\bv_k$.
\label{lem4}
\end{lemma}
\begin{IEEEproof}
First, we note that the left hand-side of \eqref{eqlem45} can be written as
\begin{align}
\frac{1}{R'}\sum_{k'\in\mathcal{R}'}\mathbb{E}\left\|\tilde\bv_{k'}-\bar\bv\right\|^2&=\frac{1}{R'}\sum_{k'\in\mathcal{R}'}\mathbb{E}\left\|\tilde\bv_{k'}-\frac{1}{R}\sum_{g\in\mathcal{R}} \bv_g\right\|^2 + \mathbb{E}\left\|\frac{1}{R}\sum_{g\in\mathcal{R}} \bv_g-\bar\bv\right\|^2 
\label{eqlem41}
\end{align}
where we have used the fact the expectation over the resampling process, which we denote by $\mathbb{E}_{R}$, of the cross terms vanishes as follows
$$\mathbb{E}_{R}\langle\tilde\bv_{k'}-\frac{1}{R_t}\sum_{g\in\mathcal{R}} \bv_g,\frac{1}{R}\sum_{g\in\mathcal{R}} \bv_g-\bar\bv\rangle =\langle\mathbb{E}_{R} \tilde\bv_{k'}-\frac{1}{R}\sum_{g\in\mathcal{R}} \bv_g,\frac{1}{R}\sum_{g\in\mathcal{R}} \bv_g-\bar\bv\rangle= 0$$  
where the last equality follows from \eqref{res1}.
Applying Lemma \ref{lem3} to the first term in the right hand-side of \eqref{eqlem41} yields 
\begin{align}
\frac{1}{R'}\sum_{k'\in\mathcal{R}'}\mathbb{E}\left\|\tilde\bv_{k'}-\bar\bv\right\|^2&=\frac{d}{R}\sum_{k\in\mathcal{R}}\mathbb{E}\left\|\bv_k-\frac{1}{R}\sum_{g\in\mathcal{R}} \bv_g\right\|^2 + \mathbb{E}\left\|\frac{1}{R}\sum_{g\in\mathcal{R}} \bv_g-\bar\bv\right\|^2 
\label{eqlem42}
\end{align}
Let us now rewrite the first term of \eqref{eqlem42} as follows
\begin{align}
&\frac{d}{R}\sum_{k\in\mathcal{R}}\mathbb{E}\left\|\bv_k-\frac{1}{R}\sum_{g\in\mathcal{R}} \bv_g\right\|^2\\ &=\frac{d}{R}\sum_{k\in\mathcal{R}}\mathbb{E}\left\|(\bv_k-\mathbb E \bv_k ) + \left (\mathbb E \bv_k -\frac{1}{R}\sum_{g\in\mathcal{R}} \mathbb E \bv_g\right)+\left (\frac{1}{R}\sum_{g\in\mathcal{R}} \mathbb E \bv_g-\frac{1}{R}\sum_{g\in\mathcal{R}} \bv_g\right) \right\|^2  \\
&=\frac{d}{R}\sum_{k\in\mathcal{R}}\mathbb{E}\left\|\bv_k-\mathbb E \bv_k\right\|^2+\frac{d}{R}\sum_{k\in\mathcal{R}}\left\| \mathbb E \bv_k-\frac{1}{R}\sum_{g\in\mathcal{R}} \mathbb E \bv_g\right\|^2+d\mathbb E\left\|\frac{1}{R}\sum_{g\in\mathcal{R}}  \bv_g-\frac{1}{R}\sum_{g\in\mathcal{R}} \mathbb E\bv_g\right\|^2\label{eqlem43} \\&-2 \frac{d}{R} \sum_{k\in\mathcal{R}}\mathbb{E}\left\langle \bv_k-\mathbb E \bv_k ,\frac{1}{R}\sum_{g\in\mathcal{R}}  \bv_g-\frac{1}{R}\sum_{g\in\mathcal{R}} \mathbb E\bv_g \right\rangle\nonumber
\end{align}
where the remaining cross terms are obviously equal to 0. Moreover, it can be easily seen that
\begin{align}
 \frac{1}{R}\sum_{k\in\mathcal{R}}\mathbb{E}\left\langle \bv_k-\mathbb E \bv_k ,\frac{1}{R}\sum_{g\in\mathcal{R}}  \bv_g-\frac{1}{R}\sum_{g\in\mathcal{R}} \mathbb E\bv_g \right\rangle = \mathbb E\left\|\frac{1}{R}\sum_{g\in\mathcal{R}}  \bv_g-\bar\bv\right\|^2
\end{align}
which also can be written as 
\begin{align}
\mathbb E\left\|\frac{1}{R}\sum_{g\in\mathcal{R}}  \bv_g-\bar\bv\right\|^2 
=\mathbb E\left\|\frac{1}{R}\sum_{g\in\mathcal{R}}  (\bv_g- \mathbb E\bv_g) \right\|^2 =\frac{1}{R^2}\sum_{g\in\mathcal{R}}  \mathbb E \left\|\bv_g- \mathbb E\bv_g\right\|^2
\label{e201}
\end{align}
where we note here that the cross terms vanish due to the independence between the vectors $\{\bv_g\}$.
Thus,
\begin{align}
\frac{d}{R}\sum_{k\in\mathcal{R}}\mathbb{E}\left\|\bv_k-\frac{1}{R}\sum_{g\in\mathcal{R}} \bv_g\right\|^2=\left(d -\frac{d}{R}\right)\frac{1}{R}\sum_{k\in\mathcal{R}}\mathbb{E}\left\|\bv_k-\mathbb E \bv_k\right\|^2+\frac{d}{R}\sum_{k\in\mathcal{R}}\left\| \mathbb E \bv_k -\bar \bv\right\|^2.
\label{e200}
\end{align}
Combining \eqref{e200}, \eqref{e201}  and  \eqref{eqlem42} yields the desired result in \eqref{eqlem45}.
\end{IEEEproof}

Compared to the proof of Theorem \ref{thm1}, the main difference is the term $\bw_{t+1}-\bw_t$ that appears in the right hand-side of \eqref{eq:102}, which can be expressed as
$$
\bw_{t+1}-\bw_t =  \gm(\{\tilde\bu_g^t\}_{g=1}^G) 
$$
where $\{\tilde\bu_g^t\}$ are generated from $\{\bu_g^t\}$ using the resampling method. We start by bounding the last term in \eqref{eq:102}
\begin{align}
& \mathbb{E}\|\bw_{t+1}-\bw_t+\eta f'(\bw_t)\|^2  =  \mathbb{E}\|\gm(\{ \tilde\bu_g^t\}_{g=1}^G)+\eta f'(\bw_t)\|^2\nonumber  \\&
 =  \mathbb{E}\|\gm(\{\tilde\bu_g^t+\eta f'(\bw_t)\}_{g=1}^G)\|^2
  \leq \frac{C_{s\alpha}^2 }{|R_t'|}{\sum_{g \in \mathcal R_t'}\mathbb{E}\left\| \tilde \bu_{g}^t + \eta f'(\bw_t) \right\|^2},\label{eqth2}
 \end{align}
 where the last inequality follows from Lemma \ref{ineq_gm_Byzantine}. Define $\overline\bu^t = \frac{1}{R_t}\sum_{g\in\mathcal R_t}\ex\bu_g^t$. The vectors $\bu_g^t$ are composed of two terms as follows
 $$
 \bu_g^t=  -\frac{1}{m} \sum_{n\in \mathcal{G}_{t,g}} \eta f_{n,i_{n}^t}'(\bw_{t}) + \bz_{t,g} = \bv_g^t+ \bz_{t,g}.
 $$
 where 
 $$
 \bv_g^t = -\frac{1}{m} \sum_{n\in \mathcal{G}_{t,g}} \eta f_{n,i_{n}^t}'(\bw_{t}).
 $$
 The randomness in $\bv_g^t$ is with respect to the client random selection and the stochastic gradients, while the randomness in $ \bz_{t,g}$ is with respect to the channel noise. The expectation of the random vectors $\bv_g^t$ for $g\in \mathcal{R}_t$ over the random selection of the clients, using Lemma \ref{lem3}, can be written as 
 $$
 \ex_{S}\bv_g^t = -\frac{1}{R}\sum_{n\in\mathcal{R}}  \eta f_{n,i_{n}^t}'(\bw_{t}),
 $$
where $ \ex_{S}$ denote the expectation over the clients random selection. Then, taking the expectation over the stochastic gradients, 
  $$
 \ex\bv_g^t = -\frac{1}{R}\sum_{n\in\mathcal{R}}  \eta f_{n}'(\bw_{t})
 $$
 which implies 
 $$
 \ex\bu_g^t = -\frac{1}{R}\sum_{n\in\mathcal{R}}  \eta f_{n}'(\bw_{t}) = -\eta f'(\bw_t),
 $$
 and thus
 $$
 \overline\bu^t = -\eta f'(\bw_t).
 $$
Applying Lemma \ref{lem4}, we get 
\begin{align}
&\frac{1 }{|R_t'|}\sum_{g \in \mathcal R_t'}\mathbb{E}\left\| \tilde \bu_{g}^t + \eta f'(\bw_t) \right\|^2 = \frac{1 }{|R_t'|}\sum_{g \in \mathcal R_t'}\mathbb{E}\left\| \tilde \bu_{g}^t  -\overline\bu^t  \right\|^2 \\ &=\left(d+ \frac{1-d}{R_t} \right)\frac{1}{R_t} \sum_{g\in \mathcal{R}_t}\mathbb{E} \| \bu_g^t- \mathbb{E}\bu_g^t  \|^2 +\frac{d}{R_t} \sum_{g\in \mathcal{R}_t }\norm{ \mathbb{E}  \bu_g^t -\overline\bu^t   }^2  \\ &=\left(d+ \frac{1-d}{R_t} \right)\frac{1}{R_t} \sum_{g\in \mathcal{R}_t}\mathbb{E} \| \bu_g^t + \eta f'(\bw_t) \|^2,
\label{lem4eq44}
 \end{align}
where the last result is obtained by noting that $\mathbb{E}\bu_g^t= \overline\bu^t = -\eta f'(\bw_t)$.
From the proof of Theorem \ref{thm1}, it holds that 
\begin{align}
\frac{1 }{R_t}{\sum_{g\in \mathcal R_t}\mathbb{E}\left\| \bu_{g}^t + \eta f'(\bw_t) \right\|^2}\leq  \eta^2\! \left(\delta^2+\kappa^2+\frac{p\sigma^2}{mPh_{min}^2} K^2\right).
\end{align}
Thus, 
\begin{align}
\frac{1 }{|R_t'|}\sum_{g \in \mathcal R_t'}\mathbb{E}\left\| \tilde \bu_{g}^t + \eta f'(\bw_t) \right\|^2 & \leq  \left(d+ \frac{1-d}{R_t} \right) \eta^2 \left( \delta^2 +\kappa^2 +  \frac{p\sigma^2}{m P h_{min}^2} K^2\right).
\label{eqth3}
 \end{align}
Combining \eqref{eqth3} with \eqref{eqth2} and noting that $R_t\geq G-B$, it holds that
\begin{align}
 \mathbb{E}\|\bw_{t+1}-\bw_t+\eta f'(\bw_t)\|^2 \!\leq C_{s\alpha}^2 \eta^2\! \left(d+ \frac{1-d}{G-B}\right)\left(\delta^2+\kappa^2+\frac{p\sigma^2}{m P h_{min}^2} K^2\right)\!.
 \label{eq202}
\end{align}
Combining \eqref{eq:102} and \eqref{eq202} yields for $\eta< \min(\frac{\mu}{2L^2} , \frac{2}{\mu})$
$$
\delta_{t+1}\leq  ( 1-\eta\mu)   \delta_t +\eta \mu A_2,
$$
where
$
A_2  \triangleq \frac{2}{\mu^2}C_{s\alpha}^2  \left(d+ \frac{1-d}{G-B}\right)\left(\delta^2+\kappa^2+\frac{p\sigma^2}{mPh_{min}^2} K^2\right)
$. Thus,
\begin{align}
\delta_{t+1} & \leq ( 1-\eta\mu)^{t+1} \delta_0 + \eta \mu A_2\sum_{i=0}^{t}( 1-\eta\mu)^{i}\\
& =  ( 1-\eta\mu)^{t+1} ( \delta_0 - A_2)+A_2,
\end{align}
which completes the proof.

\bibliographystyle{IEEEtran}
\bibliography{IEEEabrv,IEEEconf,../references}
\end{document}